\documentclass[10pt,twocolumn,letterpaper]{article}

\usepackage[pagenumbers]{cvpr} 

\usepackage{graphicx}
\usepackage{amsmath}
\usepackage{amssymb}
\usepackage{booktabs}
\usepackage{color}
\usepackage{xcolor}
\usepackage{xspace}

\newcommand{\method}{CodedVTR\xspace}

\usepackage{bm}
\usepackage{multirow}

\usepackage[pagebackref,breaklinks,colorlinks]{hyperref}

\usepackage[capitalize]{cleveref}
\crefname{section}{Sec.}{Secs.}
\Crefname{section}{Section}{Sections}
\Crefname{table}{Table}{Tables}
\crefname{table}{Tab.}{Tabs.}

\def\cvprPaperID{1839} 
\def\confName{CVPR}
\def\confYear{2022}

\begin{document}

\title{\method: Codebook-based   \\
Sparse Voxel Transformer with Geometric Guidance}

\author{Tianchen Zhao$^1$, Niansong Zhang$^1$, Xuefei Ning$^1$, He Wang$^2$, Li Yi$^{13\thanks{Corresponding Author}}$  , Yu Wang$^1$ \\
$^1$Tsinghua University \quad $^2$Peking University \quad $^3$Shanghai Qizhi Institude\\
\tt\small \{suozhang1998,foxdoraame,ericyi0124\}@gmail.com \quad niansong.zhang@outlook.com \\ 
\tt\small hewang@pku.edu.cn \quad yu-wang@mail.tsinghua.edu.cn }






\maketitle

\begin{abstract}

Transformers have gained much attention by outperforming convolutional neural networks in many 2D vision tasks. 
However, they are known to have generalization problems and rely on massive-scale pre-training and sophisticated training techniques.   
When applying to 3D tasks, the irregular data structure and limited data scale add to the difficulty of transformer's application. 
We propose \method (\textbf{Code}book-based \textbf{V}oxel \textbf{TR}ansformer), which improves data efficiency and generalization ability for 3D sparse voxel transformers.
On the one hand, we propose the codebook-based attention that projects an attention space into its subspace represented by the combination of  ``prototypes'' in a learnable codebook. It regularizes attention learning and improves generalization. 
On the other hand, we propose geometry-aware self-attention that utilizes geometric information (geometric pattern, density) to guide attention learning. \method could be embedded into existing sparse convolution-based methods, and bring consistent performance improvements for indoor and outdoor 3D semantic segmentation tasks.



\end{abstract}


\begin{figure}[t]
  \centering
    \includegraphics[width=\linewidth]{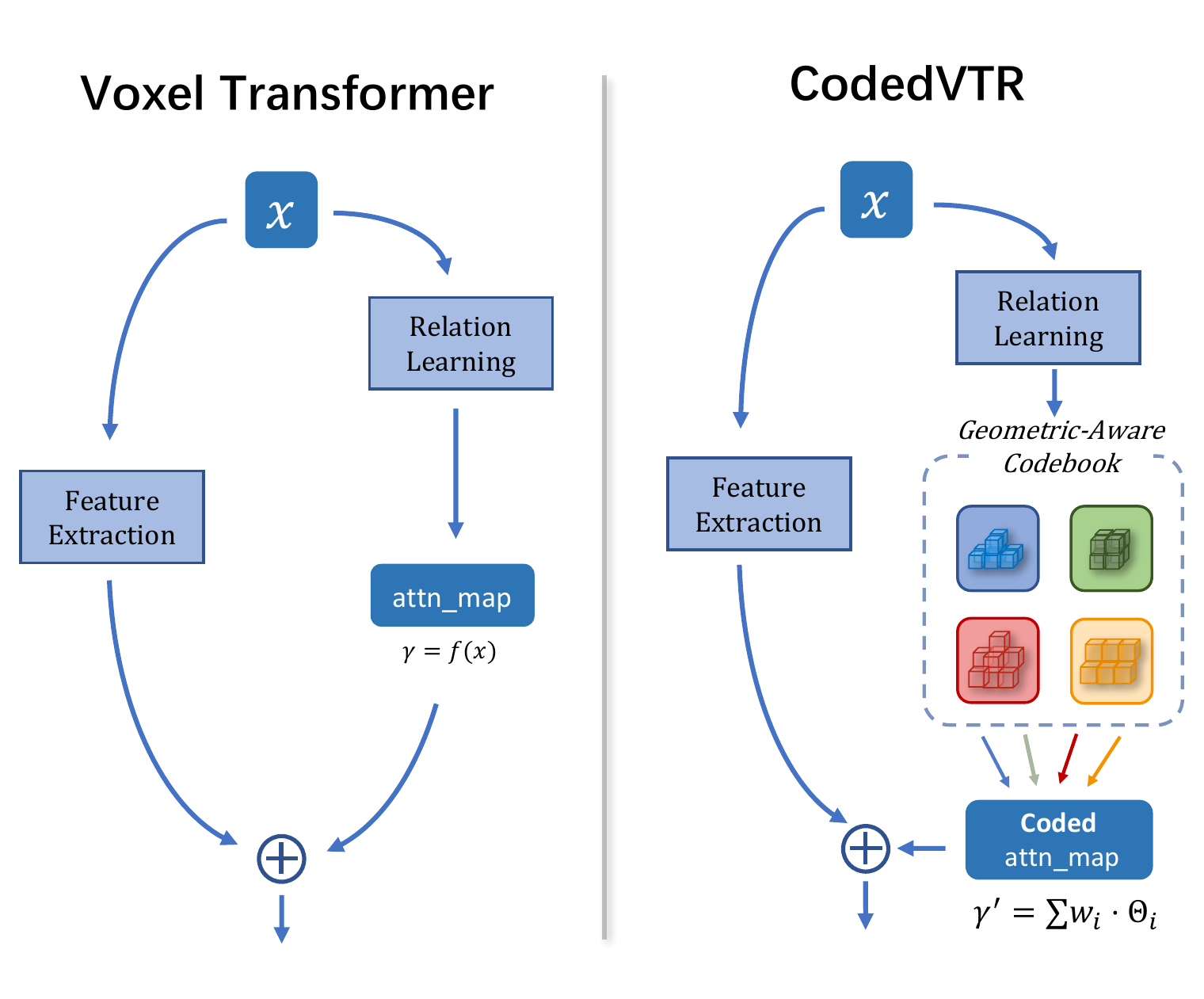}
    \caption{\textit{\method} is a novel transformer-based building block for voxel-based 3D scene understanding tasks. We propose the \textbf{codebook-based attention} to alleviate the transformer's generalization problem, which is exacerbated in 3D tasks. Taking 3D point cloud's unique properties into consideration, we also propose the \textbf{geometry-aware attention} to leverage the geometric information into attention learning, \label{fig:teaser}}
\end{figure}

\section{Introduction}
\label{sec:intro}








Recent advances in deep learning have significantly pushed forward representation learning on 3D point clouds. 3D deep learning models enable autonomous driving and robotic systems to perceive raw 3D sensor data. In this process, the 3D semantic segmentation task plays a crucial role in real-world scene understanding. It aims to classify each point into pre-defined semantic categories (e.g., car, pedestrian, table, floor, etc.), which provides point-wise perception information of the whole 3D scene. For 3D semantic segmentation on large-scale scenes,  point-based methods~\cite{kpconv,pointnet++,pointtransformer} split the scene into cubical chunks and apply point-wise operations to each chunk. Differently, voxel-based approaches~\cite{mink,cylinder,sparseconvnet,spvnas} directly voxelize the whole scene and apply 3D sparse convolution to voxels, which usually yields superior efficiency and performance. In this paper, we follow the voxel-based scheme and focus on designing sparse voxel-based architecture for the 3D semantic segmentation task. 



Transformers~\cite{transformer} have received massive attention and achieved state-of-the-art performance in many vision tasks such as image classification and semantic segmentation~\cite{swin}. It discards the convolution-like inductive bias and adopts more general-purpose self-attention operations. Although less inductive bias gives transformers potentially better representative ability, it also poses challenges to its generalization capability~\cite{battle,deit}. Current transformers rely on large-scale data pre-training (ImageNet22K), strong data augmentation, and sophisticated hyper-parameter tuning to outperform convolution neural networks (CNNs). As the ViT~\cite{vit} poses: when directly trained on the ImageNet, transformers ``yield modest accuracies of a few percentage points below ResNets of comparable size''. Prior studies~\cite{on-the-relation, can-transformer} prove that transformers have better expressive power compared with CNNs, and conclude that their poor performance on smaller datasets mainly comes from the poor generalization ability. 

This problem is further aggravated on 3D tasks. 3D point cloud data has unique properties such as sparse and irregular structure, varying density, implicit geometric features in data locality~\cite{dancenet, Mao_2019_ICCV,Khoury_2017_ICCV}. For example, 95$\%$ of the voxels that locate farther than 5m from the scene center are empty in semanticKITTI~\cite{cylinder}. Besides, different voxels have distinct local geometry patterns, e.g., the voxels on the desk and floor mainly have horizontal neighbors. Due to the large variance of density and geometric pattern in 3D data, capturing and adapting to these varying situations bring challenges for transformer's generalization. 
On the other hand, since high-quality labeled data are hard to acquire for 3D tasks, the dataset sizes are usually restricted, which also poses challenges to the generalization of transformer. 



The above-mentioned phenomena stress the vitality of overcoming the generalization problem for transformers, especially for 3D tasks. 
We seek to address the generalization problem from the perspective of architecture design.
We propose the \textbf{codebook-based attention} that regularizes the attention space to improve generalization. Specifically, we project self-attention maps into a subspace represented by the combination of several codebook elements (See Fig.~\ref{fig:teaser}). In this way, the dimension of attention learning is reduced from the whole attention space to several ``prototypes'' within the codebook. The projection constrains the attention learning to the subspace and could be viewed as a form of regularization for better generalization. 

The above codebook plays a vital role in our design. Instead of restricting the spatial support of the codebook elements to be regular cubical regions, we in addition propose \textbf{geometry-aware attention} to incorporate inductive bias regarding the spatial patterns of 3D voxel data.
Unlike pixels that have a regular and dense layout, the voxels have distinct geometric patterns and densities\footnote{We summarize the geometric pattern and density of voxels as their ``geometric information''.}. Existing sparse voxel CNNs~\cite{mink} and transformer~\cite{votr} use a fixed receptive field (spatial support) for all voxels. They solely rely on the learning process to extract geometric features without exploiting the voxel's geometric pattern. In contrast, 
we carefully design geometric regions of different shapes and ranges by collecting and clustering the local sparse patterns in the input voxel data. Taking advantage of the codebook design, we assign these geometric regions to each codebook element as its attention spatial support. In this way, we could encourage attention learning to be adaptive to the sparse pattern of voxels. 




The contributions of this work could be summarized into three aspects:

1) We address the transformer's generalization problem for the 3D domain, and propose \method, a transformer-based 3D backbone. The codebook-based self-attention projects the attention space into its subspace, serving as a regularization for better generalization.


2) We propose geometry-aware attention that exploits the sparse patterns of 3D voxel. It incorporates the inductive bias of geometric properties to guide the attention learning.


3) Our \method block can be effortlessly embedded into existing sparse convolution-based backbones. Replacing the sparse convolution with our \method brings consistent performance improvement on both indoor and outdoor 3D semantic segmentation datasets. 

\section{Related Works}
\label{sec:rw}

\subsection{Voxel-based 3D Semantic Segmentation}
\label{sec:rw-voxel}

Deep learning for 3D scene understanding has gained much popularity in recent years. 3D semantic segmentation task is a representative task, in which a semantic label is assigned to every point in a given 3D point cloud. Earlier point-based approaches like~\cite{pointnet++,kpconv} split the scene into cubical chunks and process points in each chunk. These approaches struggle to yield satisfying performance on large-scale scenes like ScanNet~\cite{scannet} or SemanticKITTI~\cite{semantic_kitti}. Alternatively, voxel-based methods like MinkowskiNets~\cite{mink} and SparseConvNets~\cite{sparseconvnet} apply the 3D voxelization to convert the irregular point cloud into regular sparse voxels to apply 3D convolution. They could directly process the whole scene and yield state-of-the-art performance on large-scale scene datasets. However, due to  uniform-voxelization and restricted receptive field, sparse convolution struggles to handle the varying density and long-term relation modeling. In this paper, we seek to improve the voxel-based feature extractor with the more flexible transformer architecture, and propose geometry-aware self-attention to address the above-mentioned problems.  

\subsection{Transformer and Self-attention}
\label{sec:rw-tr}

Transformers have achieved great success in both NLP~\cite{transformer} and vision~\cite{vit} tasks. Performance and theoretical analysis show that the self-attention operation has better expressive ability than convolution for its weaker inductive bias. Cordonnier et al.~\cite{on-the-relation}  proves that with sufficient number of heads, the self-attention could approximate the convolution. It shows that the self-attention has better expressive ability compared with convolution, thus having the potential of being a better feature extractor. Moreover, transformers have more flexible receptive field compared with convolution and could leverage richer context information.

Some prior studies attempt to introduce the transformer to 3D point clouds. For point-based methods,  PointTR~\cite{pointtr} reformulates the point cloud completion task into a set-to-set translation problem and designs transformer to solve it. Point Transformer~\cite{pointtransformer} and Pointformer~\cite{pointformer} design self-attention operator for point cloud and improve point-based methods~\cite{pointnet++} on a variety of tasks. As for voxel-based methods, VoTR~\cite{votr} designs a voxel-based transformer backbone for the 3D detection task. However, it solely adapts the self-attention in 2D vision to sparse voxels without addressing the generalization problem and 3D data's unique properties. Their reported performance and our reproduced results show that it suffers from the generalization problem. In \method, we propose \textbf{codebook-based self-attention} to alleviate such generalization problem and \textbf{geometry-aware self-attention} to leverage the geometric information for attention learning. 

\subsection{Generalization Issue of Transformer}
\label{sec:rw-tr-train}

The better representative ability of Transformer comes from its weaker inductive bias. 
However, it also makes transformers suffer from generalization problems and perform poorly on smaller-sized datasets~\cite{can-transformer, deit}. For instance, ViTs could not outperform ResNets when directly trained on CIFAR-10 or even ImageNet-1K. 
Large-scale pre-training or sophisticated augmentation and hyper-parameter tuning are required for alleviating the generalization problem. Many research attempts to alleviate this problem with less resource-consuming approaches. SAM~\cite{sam} proposes a sharpness-aware optimizer to prevent the optimization from being trapped in sharp local minimal. Geng et al.~\cite{ham} regularizes the optimization with matrix decomposition. Another line of studies~\cite{convvit,cvt,earlier} augment the model's inductive bias and improve the generalization via fusing the transformer with CNN.



\section{Methods}
\label{sec:method}

In this section, we present Codebook-based Voxel Transformer (\method), a transformer-based 3D backbone with codebook-based and geometry-aware self-attention. Our design of the codebook-based and geometry-aware self-attention can be effortlessly embedded into existing 3D backbones to replace the vanilla sparse convolution. It alleviates the transformer's well-known generalization problem and works as a general building block with better performance and parameter efficiency than 3D sparse convolution. 



This section is organized as follows. In Sec.~\ref{sec:method-voxel-tr} we introduce how to adapt the self-attention design to sparse voxels and formulate the basic ``voxel transformer'' model. Then, we present our design of the codebook-based self-attention that alleviates the generalization issue of transformer in Sec.~\ref{sec:method-codebok}. 
Finally, in Sec.~\ref{sec:method-geo} we introduce the geometry-aware attention mechanism that utilizes the geometric information to guide the attention learning, which also fits well with our codebook design.

\begin{figure*}
  \centering
    \includegraphics[width=\linewidth]{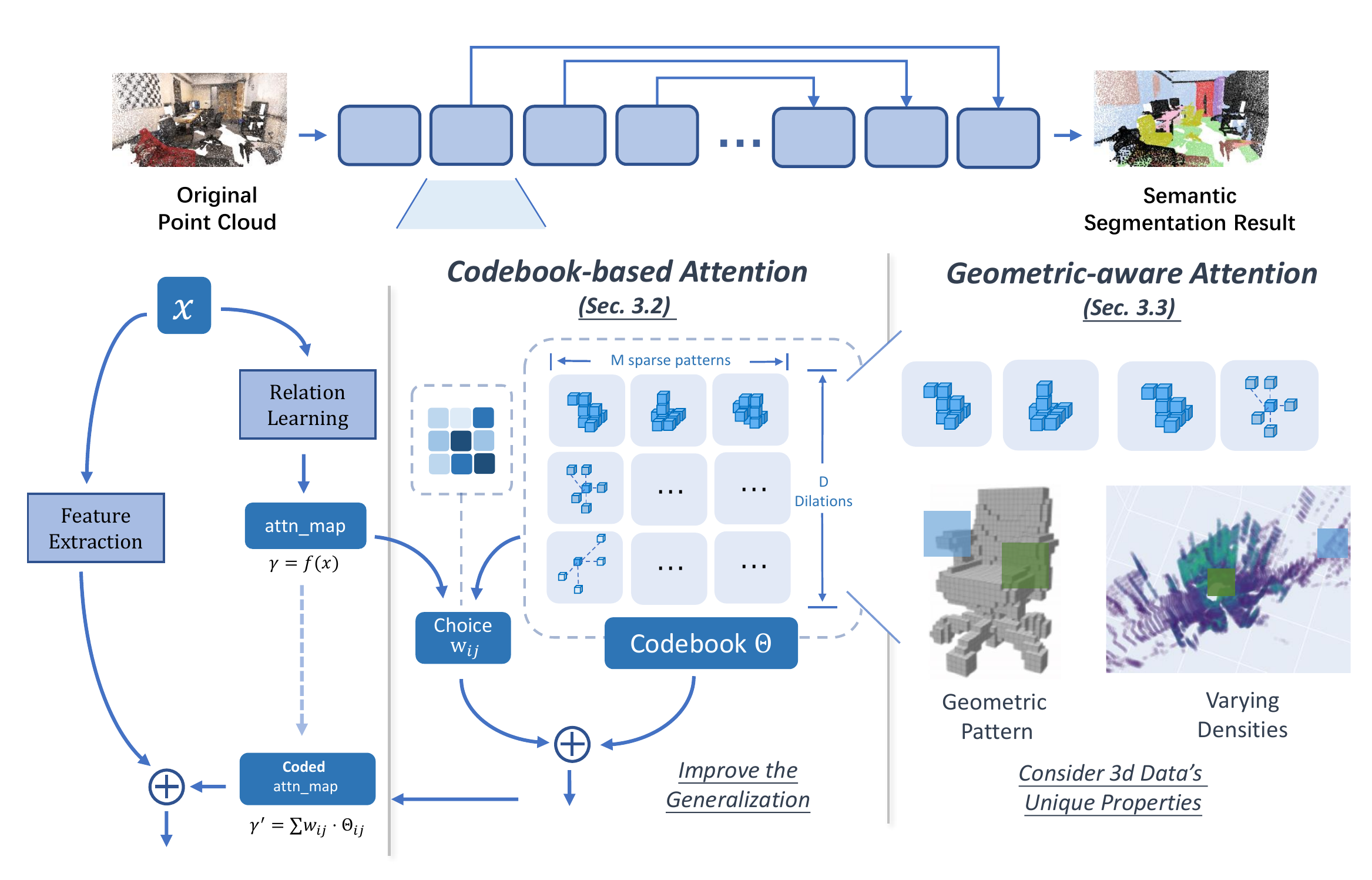}
    \caption{\textbf{Illustration of the \method block.} 
    The codebook-based attention conduct optimization in attention weight subspace represented by the codebook. The geometry-aware attention explotits the unique properties of 3D point cloud and assign various geometric regions as the attention spatial supports of codebook elements.} 
    \label{fig:flow}
\end{figure*}

\subsection{Voxel Transformer}  
\label{sec:method-voxel-tr}




For self-attention in the sparse voxel domain, we define that the point cloud is voxelized into $N$ sparse voxels with coordinates $p \in \mathbb{R}^{N \times 3}$ and features $x \in \mathbb{R}^{N \times C}$. To calculate the output for a voxel $x_i$, we conduct the self-attention operation in the local region $\varphi(x_i)$. The neighborhood $\varphi(x_i)$ represents the voxels located at nearby locations with some coordinate offsets (e.g., $(+1,-1,+1),...,(0,-1,+1)$) from the centroid $x_i$. As discussed in prior studies~\cite{pointtransformer, vec-attn}, self-attention could be decomposed into two parts: the ``relation learning'' part $\bm{f}$  and the ``feature extraction'' part $\bm{g}$. The ``relation learning'' part models the relative relations between the input voxel and its neighbors and generates the attention map $\bm{f}(\varphi(x_i))$. The attention map is then 
multiplied with the feature part and aggregated to produce the output. 
The calculation of the corresponding output $y_i$ of the input voxel $x_i$ goes as follows:
\begin{equation}
  \begin{split}
  y_i &= 
  \sum_{x_j \in \varphi(x_i)} f_{ij}(\varphi(x_i)) \odot g(x_j)
  \end{split}
  \label{equ:attn}
\end{equation}
where $f_{ij}(\varphi(x_i)) \in \mathbb{R}^{H}$ denotes the attention weight from voxel $x_j$ to $x_i$ (H is the head number), and $g(x_j)$ is the value at voxel $x_j$. 
For notation simplicity, we use a vector instead of a matrix to represent the attention map $\bm{f}(\varphi(x_i)) = \mbox{Concat}(\{f_{ij}(\varphi(x_i))\}_{x_j \in \varphi(x_i)}) \in \mathbb{R}^{N(\varphi)H}$, and write it as $\bm{f}(x)$ in the following texts, where $N(\varphi)$ stands for the neighbor size.




\subsection{Codebook-based Self-attention}
\label{sec:method-codebok}

The attention map $\bm{f}(x)$ for a certain voxel $x$ represents its relation with its $N(\varphi)$ neighbors. 
And we propose to constrain this attention to a learnable subspace through projection. Specifically, the projected attention is computed as the combination of multiple elements in a codebook. These codebook elements can be viewed as the ``prototypes'' of the attention weights and are jointly learned during the training process.




Fig.~\ref{fig:flow} gives out the graphical illustration of the codebook-based self-attention.
To be more specific, 
the attention subspace is represented using the codebook elements $\Theta=\{\theta_1, \cdots, \theta_{K}\} \in \mathbb{R}^{K \times  N(\varphi) H}$, where $K$ denotes the codebook size and $H$ denotes the head number. And the original attention $\bm{f}(x)  \in \mathbb{R}^{N(\varphi) H}$ is projected onto this subspace to become $\bm{f}_p(x) \in \mathbb{R}^{N(\varphi) H}$, where the subscript $p$ denotes ``projection''. The projection goes as follows:

\begin{equation}
  \begin{split}
  & \bm{w} = \sigma(\{\psi(\theta_{i}, \bm{f}(x)\}_{i=1,\cdots,K}) \in \mathbb{R}^{K}\\
  & \bm{f}_p(x) = \Theta^T \bm{w}
  \end{split}
  \label{equ:codebook}
\end{equation}
where $\psi$ calculates the similarity between the original attention $\bm{f}(x)$ and the codebook elements $\{\theta_i\}_{i=1,\cdots,K}$. $\sigma$ denotes the softmax operation, and $\bm{w}$ is a ``soft'' choice of the codebook elements.

Our experimental results verify the effectiveness of the codebook-based self-attention. And its performance gain can be understood from two aspects. On the one hand, the subspace projection of attention can be viewed as a type of regularization that improves the generalization ability of the transformer. On the other hand, the codebook-based self-attention can be viewed as an intermediate design that bridges the architecture gap between convolution and vanilla self-attention. The two extreme cases of the codebook-based self-attention represent convolution and vanilla self-attention, respectively. When the codebook has only one element ($K=1$), it works like a convolution block. Whereas when $K$ is sufficiently large, the subspace spanned by the codebook elements can easily have full rank. In this case, the codebook-based self-attention only conducts a mapping into a full-rank subspace without dimension reduction, and is similar to the vanilla self-attention. In another word, our design has a hyper-parameter (the codebook size $K$) that can flexibly adjust the trade-off between the generalization difficulty and representation capacity.

\subsection{Geometry-aware Self-attention}
\label{sec:method-geo}




Unlike the 2D pixels that have a dense and regular layout, the 3D voxels are sparse and irregular. Our geometry-aware attention design takes these two unique properties of the 3D sparse voxel into consideration, i.e., the sparse geometric pattern and the varying density. 

\begin{figure}[t]
  \centering
    \includegraphics[width=\linewidth]{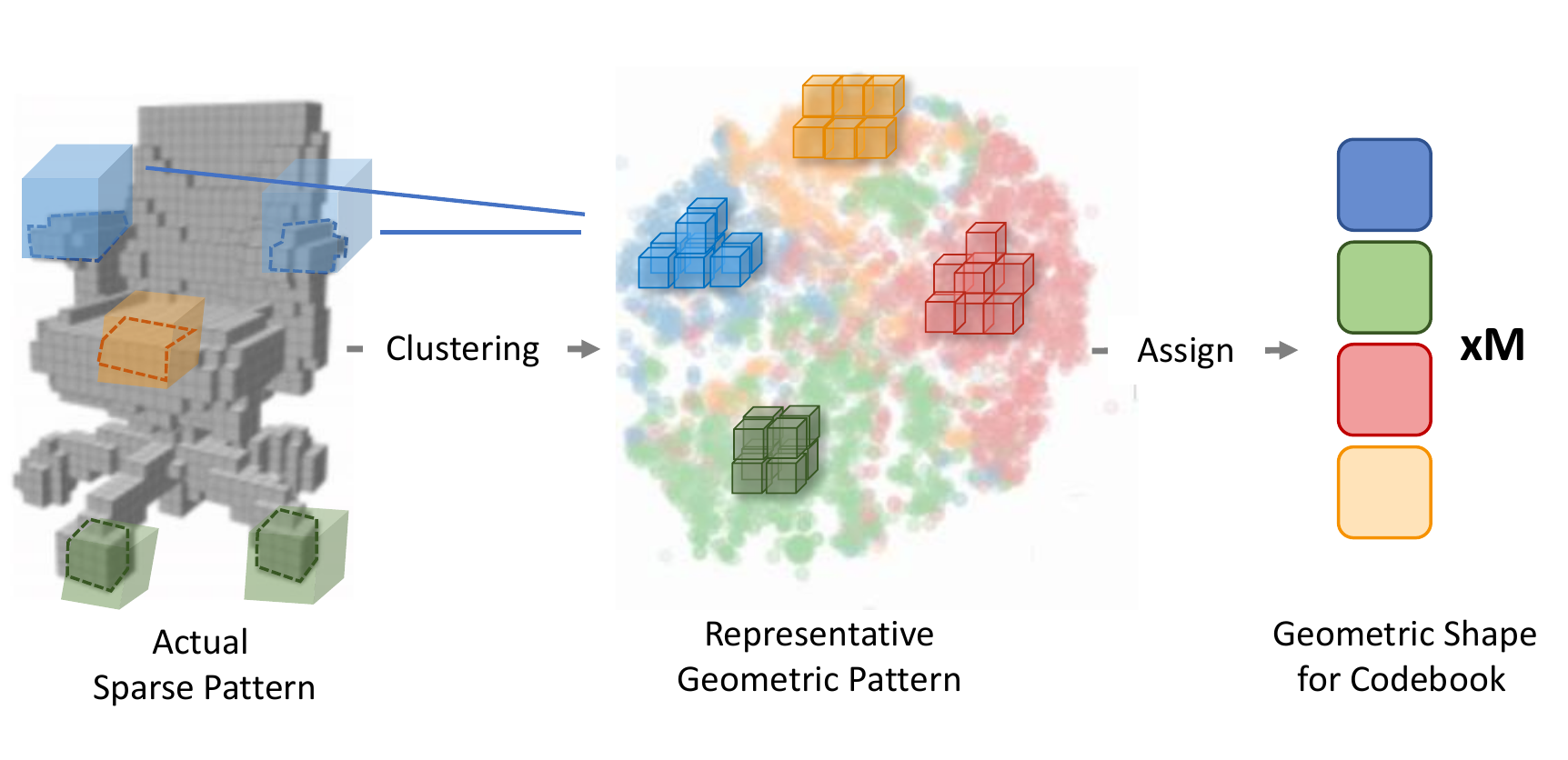}
    \caption{\textbf{Illustration of the design of geometric regions.} We conduct clustering on sparse patterns, and generate representative geometric pattern. The clustering is conducted respectively for different dataset/stride/dilation. These geometric regions are then adopted as the attention spatial support for each codebook element.  
    \label{fig:geo}}
\end{figure}

Specifically, We design various geometric regions with different shapes and ranges (as shown in the Fig.~\ref{fig:flow}). Then we assign them to each codebook element as its attention spatial support. 
These geometric regions are made up by combining $M$ distinct shapes and $D$ different dilations ($M\times D$ types of attention spatial supports in total), corresponding to the varying geometric patterns and densities of 3D data respectively. And our model can learn to adapt the weights of these codebook elements with varying attention spatial supports.



\paragraph{M distinct shapes} As illustrated in Fig.~\ref{fig:geo}, we acquire $M$ representative sparse patterns by applying the K-mode~\cite{kmodes,k-prototypes} clustering algorithm on sparse patterns. The clustering generates $M$ centroids, and we choose them as the shapes of geometric regions. This process could be interpreted as ``decomposing'' the full cubic region to the combination of a few representative sparse patterns. Instead of using a single cubic kernel for all voxels with different geometric patterns, we decouple the parameter learning of different geometric patterns and make the attention adaptive to the sparse pattern of voxels. Notice modern semantic segmentation networks usually consist of spatial pooling to quickly increase the receptive field, resulting varying spatial resolutions. We conduct clustering at each spatial resolution to obtain the respective geometric patterns.

\paragraph{D different dilations}

We adopt the above-mentioned clustering on $D$ dilations and obtain $M \times D$ regions of different shapes and ranges. The flexible ranges are critical for handling the varying density issue. If the attention feature and relation learning is constrained to a fixed local region, the voxels in low density regions may have few neighboring voxels and fail to aggregate features. After introducing different dilations, these voxels have the chance to choose the geometric region from a longer range for proper neighborhood aggregation. 



\paragraph{Computing the geometry-aware self-attention}

Aside from assigning various geometric regions and letting the attention discover geometry-aware features on its own, we introduce explicit ``geometric guidance'' for attention learning, which can be described with the following equation:

\begin{equation}
  \begin{split}
    \xi_{ij} = \cap(\varphi&(x), \varphi(\theta_{ij}))/N(\varphi(\theta_{ij})) \\
    \bm{w}^{'} =& \sigma(\{\sum^{D}_{j} \xi_{ij}/T\}_{i=1,...,M})\\
        &\otimes \sigma (\{\sum^{M}_{i} \xi_{ij}/T\}_{j=1,...,D}) \in \mathbb{R}^{M \times D} \\
    \bm{w_{f}}& = \bm{w} \odot \bm{w^{'}} \\
  \end{split}
  \label{equ:geo-guide}
\end{equation}
where $\theta_{ij}$ denotes the codebook element with the i-th geometric shape and the j-th dilation. To be more specific, we use a variable $\xi$ to describe the relative ``matching degree'' between the geometric region $\varphi(\theta_{ij})$ and the sparse neighborhood $\varphi(x)$ of the voxel $x$. ``$\cap$'' calculates the intersection of these two regions, and the result is normalized by the geometric region size $N(\varphi(\theta_{ij}))$ to produce $\xi$. Then we apply the softmax with temperature $T$ on $\xi$ for both dimensions and generate $\bm{w^{'}} \in \mathbb{R}^{M \times D} $ as ``geometry-aware choice''. The temperature $T$ controls the ``steepness'' of the $\bm{w^{'}}$ distribution, which stands for the ``strength'' of the explicit geometric guidance. Finally, we multiply $\bm{w} \in \mathbb{R}^{M \times D} $ ($\bm{w} \in \mathbb{R}^{K} $ is generated in Eqn.~\ref{equ:codebook} and then reshaped into $\mathbb{R}^{M \times D}$ ) with $\bm{w^{'}} \in \mathbb{R}^{M \times D}$ to generate the final ``choice'' $\bm{w_f}$. The ``geometric guidance'' 
explicitly enforces the attention to lean towards the geometric region that matches the sparse pattern of the voxel. For instance, the voxels on the floor should relate more closely with the ``plane'' shaped region, and voxels with low density tend to choose the region with larger dilation. It explicitly models the geometric pattern of voxels in space which CNN and transformer neglect, and leverage the geometric information to guide   self-attention learning.




\begin{table*}[t]
  \centering
  \begin{tabular}{|c|c|c|c|c|}
    \toprule
   Dataset  & \multicolumn{2}{|c|}{Method (Model)}  & Param & mIOU \\
    \midrule
    \multirow{6}{*}{ScanNet} & \multirow{2}{*}{Convolution} & Minkowski-M~\cite{mix3D} & 7M & 67.3\% \\
                             & & Minkowski-L~\cite{mink} & 11M & 72.4\% \\
                            \cmidrule{2-5}
                            & \multirow{5}{*}{Transformer} & PointTransformer~\cite{pointtransformer}$\dagger$ & 6M & 58.6\% \\
                            & & VoTR (Mink-M) ~\cite{votr}$\dagger$ & 7M & 62.5\% \\
                            & & VoTR (Mink-L) ~\cite{votr}$\dagger$ & 11M &  66.1\% \\
                            & & CodedVTR (Mink-M) & 7M & \textbf{68.8}\% \\
                            & & CodedVTR (Mink-L) & 11M & \textbf{73.0}\% \\
    \midrule
    \multirow{9}{*}{SemanticKITTI} & \multirow{3}{*}{Convolution} & 
                             Minkowski-M~\cite{spvnas}  & 7M & 58.9\% \\
                            & & Minkowsk-L~\cite{spvnas} & 11M & 61.1\% \\
                            & & SPVCNN~\cite{spvnas} & 8M & 60.7\% \\
                            \cmidrule{2-5}
                            & \multirow{5}{*}{Transformer} & VoTR (Mink-M)~\cite{votr} $\dagger$  & 7M & 56.5\% \\
                            & & VoTR (Mink-L)~\cite{votr}$\dagger$ & 11M & 58.2\% \\
                            & & CodedVTR (Mink-M) & 7M & \textbf{60.4}\% \\
                            & & CodedVTR (Mink-L) & 11M & \textbf{63.2}\% \\
                            & & CodedVTR (SPVCNN) & 8M & \textbf{61.8}\% \\
    \midrule
    \multirow{4}{*}{Nuscenes} & \multirow{2}{*}{Convolution} & 
                              Minkowski-M~\cite{spvnas}  & 7M &  66.5\% \\
                            & & Minkowsk-L~\cite{spvnas} & 11M & 69.4\% \\ 
                            \cmidrule{2-5}
                            & \multirow{2}{*}{Transformer} & CodedVTR (Mink-M) & 7M & \textbf{69.9}\% \\
                            & & CodedVTR (Mink-L) & 11M & \textbf{72.5}\% \\

    \bottomrule
  \end{tabular}
  \caption{3D Semantic Segmentation peformance on ScanNet and SemanticKITTI val set. ``$\dagger$'' denotes the result is reproduced due to absence in the original paper. ``M/L'' denotes the model has similar depth and width like ``MinkowskiResNet20/42''.\label{tab:results}}
\end{table*} 

\section{Experiments}
\label{sec:exp}

In this section, we evaluate our \method on the popular indoor and outdoor 3D semantic segmentation datasets:  ScanNet~\cite{scannet} and the SemanticKITTI~\cite{semantic_kitti}. We give detailed descriptions of the implementation details about our \method and present its performance on these datasets. For each dataset, we first briefly introduce the dataset and evaluation metrics. Then we compare the performance of our \method with the sparse voxel CNN and transformer and demonstrate its superior performance. We also show that \method could be embedded into existing sparse convolution-based methods and further improve their performance from the aspect of architectural design.



\subsection{Implemention Details}
\label{sec:method-detail}



For compatibility with current backbones, we adopt the U-Net-like block layout like the well-known MinkowskiNet~\cite{mink} (as the upper part of the Fig.~\ref{fig:flow}) and design a \method block to replace ResNet-like sparse convolution block. We also adapt the VoTR~\cite{votr} to our architecture as the voxel transformer baseline. In Table.~\ref{tab:results}, the ``M'' or ``L'' suffixes mark the medium-sized or large-sized models, which have similar depth and width like MinkowskiResNet-20 and MinkowskiResNet-42. ``CodedVTR-X'' denotes embedding \method into other convolution-based methods. 

We supplement some details about our \method block as follows. 
We take the feature within the local neighborhood $\varphi$ and generate the attention at the corresponding location. 
More specifically, the $\bm{f}$ in $\bm{f}(\varphi(x_i))$ consists of a linear layer and a lightweight channel-wise convolution for local aggregation within $\varphi(x_i)$ like previous attention-based methods~\cite{cbam}. The local aggregation already models the spatial information thus the positional encoding could be discarded~\cite{cvt}. 
The design of the geometric regions is conducted by applying k-modes~\cite{kmodes} clustering on the sparse patterns of voxels from 10 randomly sampled scenes within $3\times3\times3$ cubic neighborhood for each stride. The hyper-parameter $M$ is chosen through investigating the saturate point of the clustering cost function and set as 8. The above-mentioned clustering is applied on three dilations respectively, resulting in $8 \times 3$ geometric regions for the codebook. 

\subsection{Performance on the Indoor Segmentation Dataset}
\label{sec:exp-perf}

We conduct experiments on indoor 3D semantic segmentation dataset, ScanNet~\cite{scannet}, which contains various 3D indoor scenes annotated with 20 semantic classes. We follow the common setting and split 1201, 312 for train and validation respectively. We follow the training protocol of MinkowskiNet~\cite{mink} for fair comparison. As shown in Table.~\ref{tab:results}, replacing the ResNet-like sparse convolution block with our proposed \method block could bring consistent performance improvement. The middle-sized \method-M model achieves $1.5\%$ higher mIoU (68.8\%/67.3\%) with $2/3$ param size (6.1M/9.2M), the large-sized \method-L outperforms the convolution-based counterpart (+0.6\%) with half the param size (25.7M/40.2M). Besides, due to the relatively small dataset size, the voxel transformer (VoTR) suffers from severe 
generalization problem and fail to achieve comparable performance with the CNN (-6\%). It shows the effectiveness of our \method on improving the 3D transformer's generalization ability. 



\subsection{Performance on the Outdoor Segmentation Dataset}

For the outdoor scene semantic segmentation experiment, we choose to use a larger dataset, SemanticKITTI~\cite{semantic_kitti}, which consists of LIDAR scans of outdoor scenes. Among them 19130 scenes are for training, and 4071 are for validation. There are 19 semantic classes to evaluate. We also keep our training settings the same as the baseline method~\cite{spvnas} and present the results in Table~\ref{tab:results}. We observe that \method consistently outperforms both the CNN and transformer models with similar sizes (by 1.5\%/3.9\% for middle-sized model and 2.1\%/5.0\% for large-sized model). Since SemanticKITTI is a relatively larger-scale dataset than ScanNet, the generalization problem of transformer is moderate. However, it still struggles to outperform convolution (-2\%). Moreover, SPVCNN~\cite{spvnas} notably outperforms (1.8\%) Minkowski-M, and replacing the voxel branch in SPVCNN with our \method block could further boost its performance by $1.1\%$. It shows that our \method block could be easily embedded into the mainstream sparse convolution-based methods and further improve their performance from the aspect of architectural design.

\section{Analysis and Discussions}

\subsection{Analysis of the Attention Learning}
\label{sec:exp-opt}

\begin{figure}
  \centering
    \includegraphics[width=0.9\linewidth]{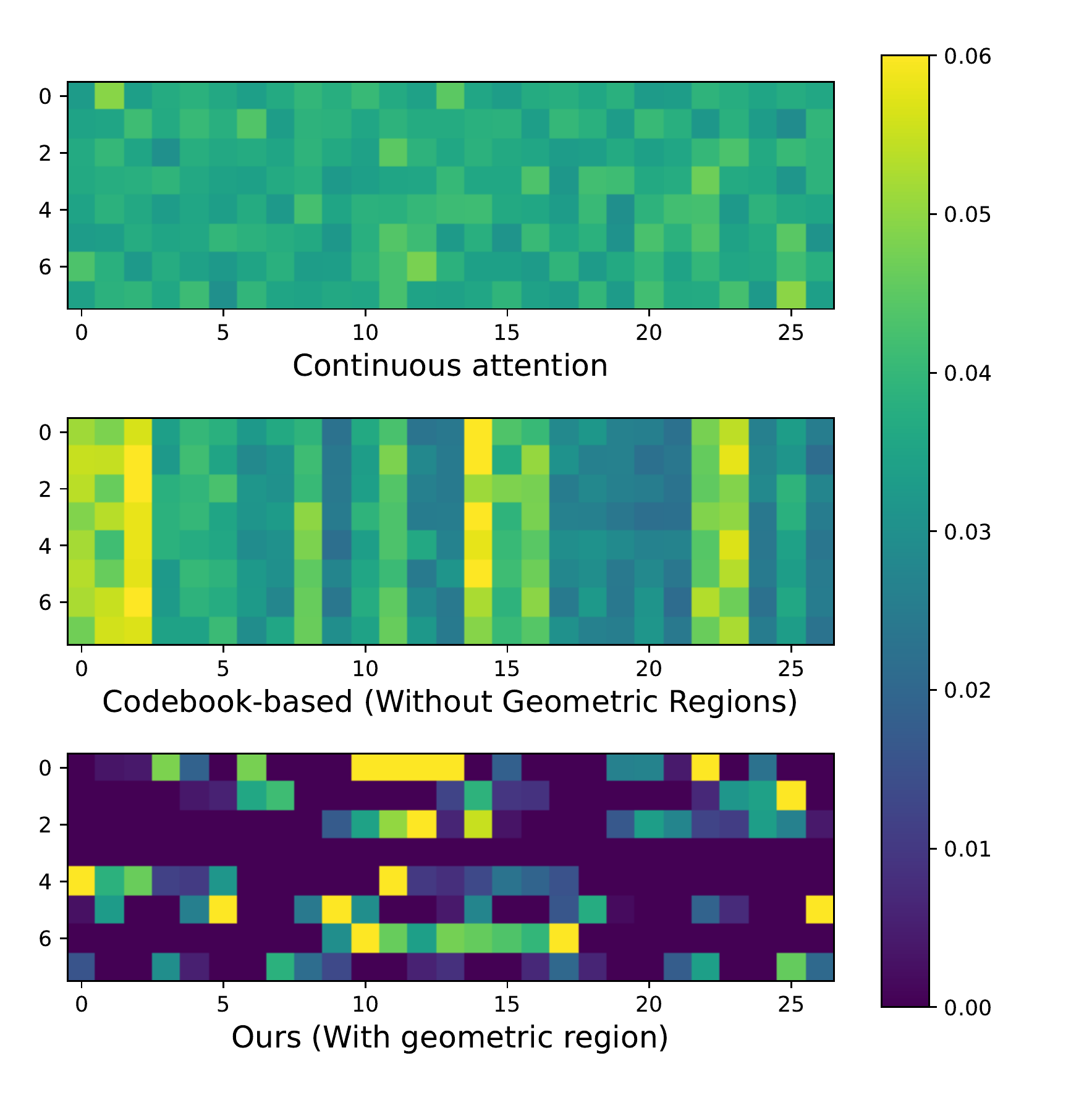}
    \caption{\textbf{Illustration of the attention maps of deeper layers for different self-attention design.} For both the ``continuous attention'' and ``codebook-based attention without geometric regions'', the attention maps collapse in latter layers (i.e., become uniform and fail to provide valuable information). }
    \label{fig:heatmap}
\end{figure}

\paragraph{Comparison with methods for transformer's generalization improvement} 
The main purpose of our \method is to alleviate the transformer's generalization problem in 3D domain. There exists many attempts for improving the transformer's generalization ability. We introduce them into the sparse voxel transformer and compare our \method with them. We compare two representative approaches from the aspects of the optimizer (SAM~\cite{sam}) and architectural design ConViT~\cite{convvit} respectively. As shown in Table.~\ref{tab:opt}, these approaches do not effectively address the problem as \method. The reason for their unsatisfactory performance is that they are designed for general-purpose vision transformer. Therefore, they did not take 3D data's unique properties into consideration, thus failing to exploit the rich geometric information to assist the attention learning. 

\paragraph{Visualization of the attention map} 
Prior literature~\cite{deepvit} points out that deeper transformers suffer from the ``attention collapse'' problem, in which the attention maps in latter layers tend to be uniform and fail to provide valuable information. We witness similar phenomenon for 3D voxel transformer, as shown in the upper heatmap of Fig.~\ref{fig:heatmap}. Introducing the codebook-based self-attention alleviates this problem to some extent. However, in deeper layers, we still witness the similar ``collapse'' problem where codebook weights grow similar and lose expressivity. After we introduce the geometry-aware attention, the ``collapse'' problem disappears. We explain this phenomenon in two ways: 1) The geometric guidance helps the attention weight learning match the distinct geometric patterns. 2) The geometric region itself  can be viewed as a form of diversity regularization on attention weights in terms of applying a hard mask on the attention weights.


\begin{table}[t]
  \centering
  \resizebox{\columnwidth}{!}{%
  \begin{tabular}{|c|c|c|}
    \toprule
    Method  & Details & mIoU \\
    \midrule
    VoxelTR$\dagger$ & -  &  56.5\% \\
    SAM~\cite{sam}$\dagger$ & Optimizer & 56.6\% \\
    ConvViT~\cite{convvit}$\dagger$ & Conv TR fusion & 57.4\% \\
    \midrule
    CodedVTR & Codebook Only &  58.1\% \\
    CodedVTR & Codebook + Geo-Region &  58.9\% \\
    CodedVTR & Codebook + Geo-Guide &  \textbf{60.4}\% \\
    \bottomrule
  \end{tabular}
  }
  \caption{\textbf{Ablation studies of \method proposed techniques and comparison of other approaches for improving transformer's generalization ability on SemanticKITTI.} ``Geo-region'' denotes that we only assign different geometric regions for codebook elements. ``Geo-guidance'' represents applying the explicit encouragement.  \label{tab:opt} }

\end{table}


\subsection{Ablation studies of \method}

\paragraph{Ablation studies of the proposed techniques} 
We conduct ablation studies in Table.~\ref{tab:opt}. The ``Codebook only'' denotes codebook-based self-attention without various geometric regions, it alleviates the generalization problem and matches the performance of convolution. However, as shown in Fig.~\ref{fig:heatmap}, the codebook-based self-attention alone could not completely address the problem. The ``Codebook+Geo-Region'' denotes that we assign distinct geometric regions to each codebook element and solely rely on attention learning. Finally, the ``Codebook+Geo-Guide'' denotes our final scheme with all the techniques. 

\paragraph{Ablation studies of the codebook design}
As we state in Sec.~\ref{sec:method-codebok}, the design of the codebook is critical for our codebook-based self-attention to strike a balance between the generalization ability and expressivity, Therefore, We also conduct ablation studies of the codebook design as shown in Table.~\ref{tab:ablation-codebook}. When the codebook has only one element, it behaves like convolution as expected. When we properly enlarge the size of $M$ and $D$, the model's expressivity and performance improves. However, too large codebook (e.g., D/M=3/16 or D/M=4/1) size burdens the optimization and causes performance degradation. Besides, we compare our learned attention with random sampling for the same codebook size. We witness significant performance differences (+6.2\%), which demonstrates the effectiveness of our attention learning.  





\begin{table}
 \centering
\begin{tabular}{|c c|c|}
\hline
\multicolumn{2}{|c|}{Codebook Design} &  \multicolumn{1}{c|}{\multirow{2}{*}{mIoU}}\\ 
\cline{1-2}
D & M & \multicolumn{1}{c|}{} \\ 
\cline{1-3}
1  & 1 &  57.1\%  \\
3  & 1 & 58.5\% \\
4  & 1 & 55.3\% \\
\hline
3 (RS)  & 8 (RS) & 54.2\% \\
3  & 8 &  \textbf{60.4}\% \\
3  & 16 &  58.1\% \\
\cline{1-3}
\end{tabular}
  \caption{Ablation study: Comparison of performance of \method-A with different codebook designs on semKITTI. Noted that the geometric regions are applied in this experiment. The ``D \& M'' stands for different dilations and sparse patterns. \label{tab:ablation-codebook}}
\end{table}

\begin{figure}
  \centering
      \vspace{10pt}
    \includegraphics[width=\linewidth]{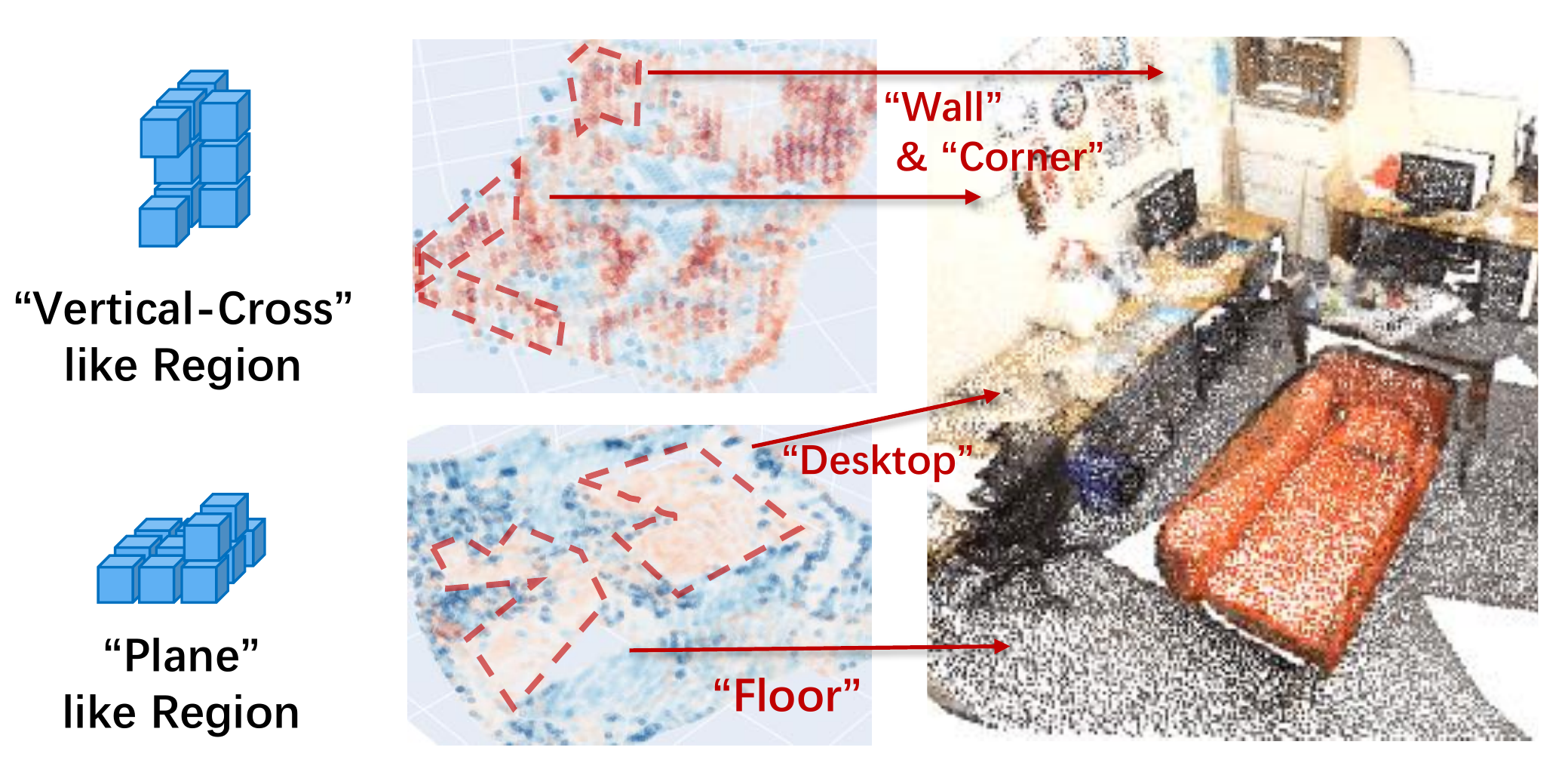}
    \caption{\textbf{Visualization of the geometry-aware self-attention's soft ``choice'' for certain geometric shape} (Red denotes larger value). The attention learns to use ``plane'' shaped region for tabletop and floor for ScanNet scene.
    \vspace{5pt}
    }
    \label{fig:sparse-pattern}
\end{figure}

\begin{figure}
  \centering

    \includegraphics[width=\linewidth]{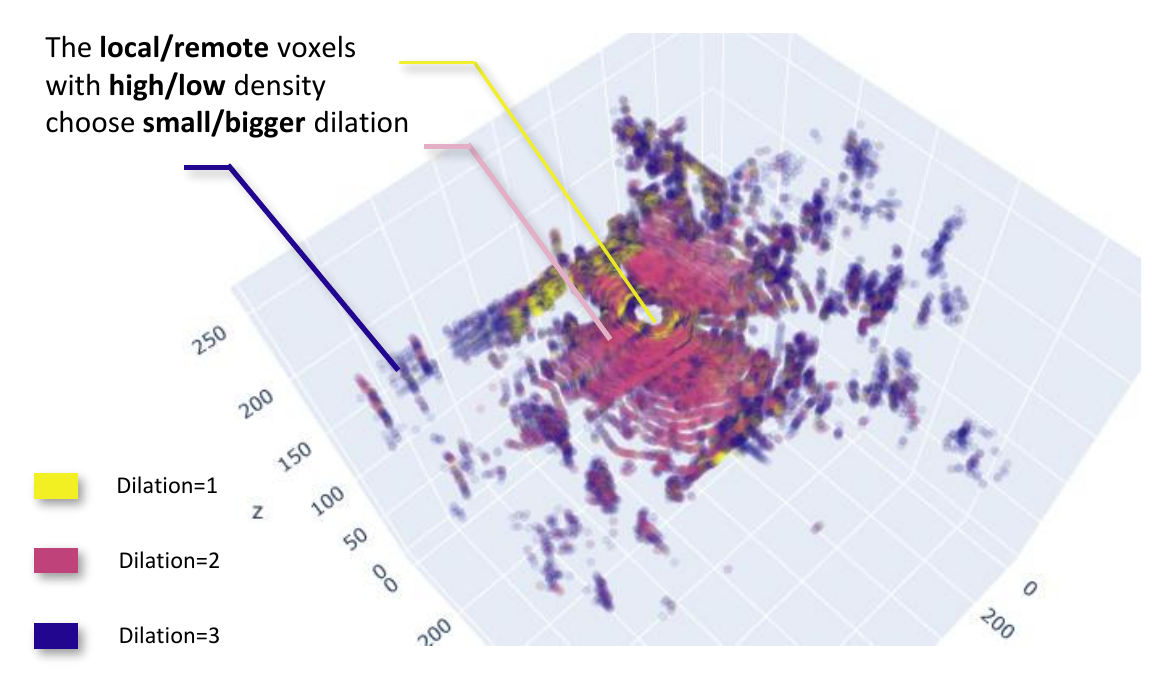}
    \caption{\textbf{Visualization of the geometry-aware self-attention's ``choice'' for different dilations.} (the color represents the best choice out of 3 dilations for certain voxel). The attention learns to use smaller dilation for local region (yellow) and larger ones for farther voxels (purple).  }
    \label{fig:density}
\end{figure}

\subsection{Visualization}
\label{sec:exp-vis}

\paragraph{\method learns to adapt to geometric patterns}
We visualize the ``choice'' value with geometric regions for a certain geometric pattern of the whole scene. As shown in Fig.~\ref{fig:sparse-pattern}, the ``plane-shaped'' geometric region relates more closely with the floor and the desktop region. Similarly, the ``vertical cross'' shaped region adapts to the wall corners. It demonstrates that the geometric regions successfully encourage the attention to adapt to the voxel's sparse pattern. 

\paragraph{\method learns to adapt to varying densities} 
We visualize the best ``choice'' along the dilation dimension of the codebook. As shown in Fig.~\ref{fig:density}, \method learns to use the yellow ones for the local region with high density, and region with larger receptive field for remote, low-density points. In this way, \method handles the varying densities and ensures proper feature aggregation for all voxels.

\section{Conclusions} 

This paper proposes \method, which seeks to address the generalization problem of transformer for 3D scene understanding. Specifically, we propose the ``codebook-based'' attention to project the attention space into a learnable subspace represented by the codebook. This technique can be seen as a a sort of regularization to improve the generalization of self-attention. Besides, we propose ``geometry-aware attention'' that assigns various geometric shapes to each codebook element and leverages the geometric feature to guide the attention learning. Extensive experiments on both indoor and outdoor 3D semantic segmentation datasets show that our \method could achieve better performance compared with both CNN and transformer. 

\section*{Acknowledgement}

This work was supported by National Natural Science Foundation of China (No. U19B2019, 61832007),
Tsinghua EE Xilinx AI Research Fund, Beijing National Research Center for Information Science and Technology (BNRist), and Beijing Innovation Center for Future Chips.

{\small
\bibliographystyle{plain}
\bibliography{egbib}

\begin{thebibliography}{10}

\bibitem{semantic_kitti}
Jens Behley, Martin Garbade, Andres Milioto, Jan Quenzel, Sven Behnke,
  C.~Stachniss, and Juergen Gall.
\newblock Semantickitti: A dataset for semantic scene understanding of lidar
  sequences.
\newblock {\em 2019 IEEE/CVF International Conference on Computer Vision
  (ICCV)}, pages 9296--9306, 2019.

\bibitem{sam}
Xiangning Chen, Cho-Jui Hsieh, and Boqing Gong.
\newblock When vision transformers outperform resnets without pretraining or
  strong data augmentations.
\newblock {\em ArXiv}, abs/2106.01548, 2021.

\bibitem{mink}
Christopher~Bongsoo Choy, JunYoung Gwak, and Silvio Savarese.
\newblock 4d spatio-temporal convnets: Minkowski convolutional neural networks.
\newblock {\em 2019 IEEE/CVF Conference on Computer Vision and Pattern
  Recognition (CVPR)}, pages 3070--3079, 2019.

\bibitem{on-the-relation}
Jean-Baptiste Cordonnier, Andreas Loukas, and Martin Jaggi.
\newblock On the relationship between self-attention and convolutional layers.
\newblock {\em ArXiv}, abs/1911.03584, 2020.

\bibitem{scannet}
Angela Dai, Angel~X. Chang, Manolis Savva, Maciej Halber, Thomas~A. Funkhouser,
  and Matthias Nie{\ss}ner.
\newblock Scannet: Richly-annotated 3d reconstructions of indoor scenes.
\newblock {\em 2017 IEEE Conference on Computer Vision and Pattern Recognition
  (CVPR)}, pages 2432--2443, 2017.

\bibitem{convvit}
St{\'e}phane d'Ascoli, Hugo Touvron, Matthew~L. Leavitt, Ari~S. Morcos, Giulio
  Biroli, and Levent Sagun.
\newblock Convit: Improving vision transformers with soft convolutional
  inductive biases.
\newblock In {\em ICML}, 2021.

\bibitem{kmodes}
Nelis~J. de~Vos.
\newblock kmodes categorical clustering library.
\newblock \url{https://github.com/nicodv/kmodes}, 2015--2021.

\bibitem{vit}
Alexey Dosovitskiy, Lucas Beyer, Alexander Kolesnikov, Dirk Weissenborn,
  Xiaohua Zhai, Thomas Unterthiner, Mostafa Dehghani, Matthias Minderer, Georg
  Heigold, Sylvain Gelly, Jakob Uszkoreit, and Neil Houlsby.
\newblock An image is worth 16x16 words: Transformers for image recognition at
  scale.
\newblock {\em ArXiv}, abs/2010.11929, 2021.

\bibitem{ham}
Zhengyang Geng, Meng-Hao Guo, Hongxu Chen, Xia Li, Ke~Wei, and Zhouchen Lin.
\newblock Is attention better than matrix decomposition?
\newblock {\em ArXiv}, abs/2109.04553, 2021.

\bibitem{sparseconvnet}
Benjamin Graham, Martin Engelcke, and Laurens van~der Maaten.
\newblock 3d semantic segmentation with submanifold sparse convolutional
  networks.
\newblock {\em 2018 IEEE/CVF Conference on Computer Vision and Pattern
  Recognition}, pages 9224--9232, 2018.

\bibitem{k-prototypes}
Zhexue Huang.
\newblock Extensions to the k-means algorithm for clustering large data sets
  with categorical values.
\newblock {\em Data mining and knowledge discovery}, 2(3):283--304, 1998.

\bibitem{Khoury_2017_ICCV}
Marc Khoury, Qian-Yi Zhou, and Vladlen Koltun.
\newblock Learning compact geometric features.
\newblock In {\em Proceedings of the IEEE International Conference on Computer
  Vision (ICCV)}, Oct 2017.

\bibitem{can-transformer}
Shanda Li, Xiangning Chen, Di~He, and Cho-Jui Hsieh.
\newblock Can vision transformers perform convolution?
\newblock {\em ArXiv}, abs/2111.01353, 2021.

\bibitem{dancenet}
Xiang Li, Lingjing Wang, Mingyang Wang, Congcong Wen, and Yi~Fang.
\newblock Dance-net: Density-aware convolution networks with context encoding
  for airborne lidar point cloud classification.
\newblock {\em ISPRS Journal of Photogrammetry and Remote Sensing},
  166:128--139, 2020.

\bibitem{swin}
Ze~Liu, Yutong Lin, Yue Cao, Han Hu, Yixuan Wei, Zheng Zhang, Stephen
  Ching-Feng Lin, and Baining Guo.
\newblock Swin transformer: Hierarchical vision transformer using shifted
  windows.
\newblock {\em ArXiv}, abs/2103.14030, 2021.

\bibitem{Mao_2019_ICCV}
Jiageng Mao, Xiaogang Wang, and Hongsheng Li.
\newblock Interpolated convolutional networks for 3d point cloud understanding.
\newblock In {\em Proceedings of the IEEE/CVF International Conference on
  Computer Vision (ICCV)}, October 2019.

\bibitem{votr}
Jiageng Mao, Yujing Xue, Minzhe Niu, Haoyue Bai, Jiashi Feng, Xiaodan Liang,
  Hang Xu, and Chunjing Xu.
\newblock Voxel transformer for 3d object detection.
\newblock {\em ArXiv}, abs/2109.02497, 2021.

\bibitem{mix3D}
Alexey Nekrasov, Jonas Schult, Or~Litany, B.~Leibe, and Francis Engelmann.
\newblock Mix3d: Out-of-context data augmentation for 3d scenes.
\newblock {\em ArXiv}, abs/2110.02210, 2021.

\bibitem{pointformer}
Xuran Pan, Zhuofan Xia, Shiji Song, Li~Erran Li, and Gao Huang.
\newblock 3d object detection with pointformer.
\newblock In {\em CVPR}, 2021.

\bibitem{pointnet++}
C.~Qi, L.~Yi, Hao Su, and Leonidas~J. Guibas.
\newblock Pointnet++: Deep hierarchical feature learning on point sets in a
  metric space.
\newblock In {\em NIPS}, 2017.

\bibitem{spvnas}
Haotian Tang, Zhijian Liu, Shengyu Zhao, Yujun Lin, Ji~Lin, Hanrui Wang, and
  Song Han.
\newblock Searching efficient 3d architectures with sparse point-voxel
  convolution.
\newblock In {\em ECCV}, 2020.

\bibitem{kpconv}
Hugues Thomas, C.~Qi, Jean-Emmanuel Deschaud, Beatriz Marcotegui, François
  Goulette, and Leonidas~J. Guibas.
\newblock Kpconv: Flexible and deformable convolution for point clouds.
\newblock {\em 2019 IEEE/CVF International Conference on Computer Vision
  (ICCV)}, pages 6410--6419, 2019.

\bibitem{deit}
Hugo Touvron, Matthieu Cord, Matthijs Douze, Francisco Massa, Alexandre
  Sablayrolles, and Herv'e J'egou.
\newblock Training data-efficient image transformers \& distillation through
  attention.
\newblock In {\em ICML}, 2021.

\bibitem{transformer}
Ashish Vaswani, Noam~M. Shazeer, Niki Parmar, Jakob Uszkoreit, Llion Jones,
  Aidan~N. Gomez, Lukasz Kaiser, and Illia Polosukhin.
\newblock Attention is all you need.
\newblock {\em ArXiv}, abs/1706.03762, 2017.

\bibitem{cbam}
Sanghyun Woo, Jongchan Park, Joon-Young Lee, and In-So Kweon.
\newblock Cbam: Convolutional block attention module.
\newblock In {\em ECCV}, 2018.

\bibitem{cvt}
Haiping Wu, Bin Xiao, Noel C.~F. Codella, Mengchen Liu, Xiyang Dai, Lu~Yuan,
  and Lei Zhang.
\newblock Cvt: Introducing convolutions to vision transformers.
\newblock {\em ArXiv}, abs/2103.15808, 2021.

\bibitem{earlier}
Tete Xiao, Mannat Singh, Eric Mintun, Trevor Darrell, Piotr Doll{\'a}r, and
  Ross~B. Girshick.
\newblock Early convolutions help transformers see better.
\newblock {\em ArXiv}, abs/2106.14881, 2021.

\bibitem{pointtr}
Xumin Yu, Yongming Rao, Ziyi Wang, Zuyan Liu, Jiwen Lu, and Jie Zhou.
\newblock Pointr: Diverse point cloud completion with geometry-aware
  transformers.
\newblock {\em ArXiv}, abs/2108.08839, 2021.

\bibitem{vec-attn}
Hengshuang Zhao, Jiaya Jia, and Vladlen Koltun.
\newblock Exploring self-attention for image recognition.
\newblock {\em 2020 IEEE/CVF Conference on Computer Vision and Pattern
  Recognition (CVPR)}, pages 10073--10082, 2020.

\bibitem{pointtransformer}
Hengshuang Zhao, Li~Jiang, Jiaya Jia, Philip H.~S. Torr, and Vladlen Koltun.
\newblock Point transformer.
\newblock {\em ArXiv}, abs/2012.09164, 2020.

\bibitem{battle}
Yucheng Zhao, Guangting Wang, Chuanxin Tang, Chong Luo, Wenjun Zeng, and
  Zhengjun Zha.
\newblock A battle of network structures: An empirical study of cnn,
  transformer, and mlp.
\newblock {\em ArXiv}, abs/2108.13002, 2021.

\bibitem{deepvit}
Daquan Zhou, Bingyi Kang, Xiaojie Jin, Linjie Yang, Xiaochen Lian, Qibin Hou,
  and Jiashi Feng.
\newblock Deepvit: Towards deeper vision transformer.
\newblock {\em ArXiv}, abs/2103.11886, 2021.

\bibitem{cylinder}
Xinge Zhu, Hui Zhou, Tai Wang, Fangzhou Hong, Yuexin Ma, Wei Li, Hongsheng Li,
  and Dahua Lin.
\newblock Cylindrical and asymmetrical 3d convolution networks for lidar
  segmentation.
\newblock In {\em Proceedings of the IEEE/CVF Conference on Computer Vision and
  Pattern Recognition}, pages 9939--9948, 2021.

\end{thebibliography}
}

\clearpage


\appendix
 
\section*{Appendix for \method: Codebook-based
Sparse Voxel Transformer with Geometric Guidance}

\section{Architecture Design}

In this section, we present the detailed model architecture. We follow the MinkowskiNet's~\cite{mink} scheme and adopt their ResNet-20 and ResNet-42 architecture, and replace their ResNet-like building block with our \method block. The \method block shares the same input and output channel size as the ResNet block.

\begin{figure}[h]
  \centering
    \includegraphics[width=\linewidth]{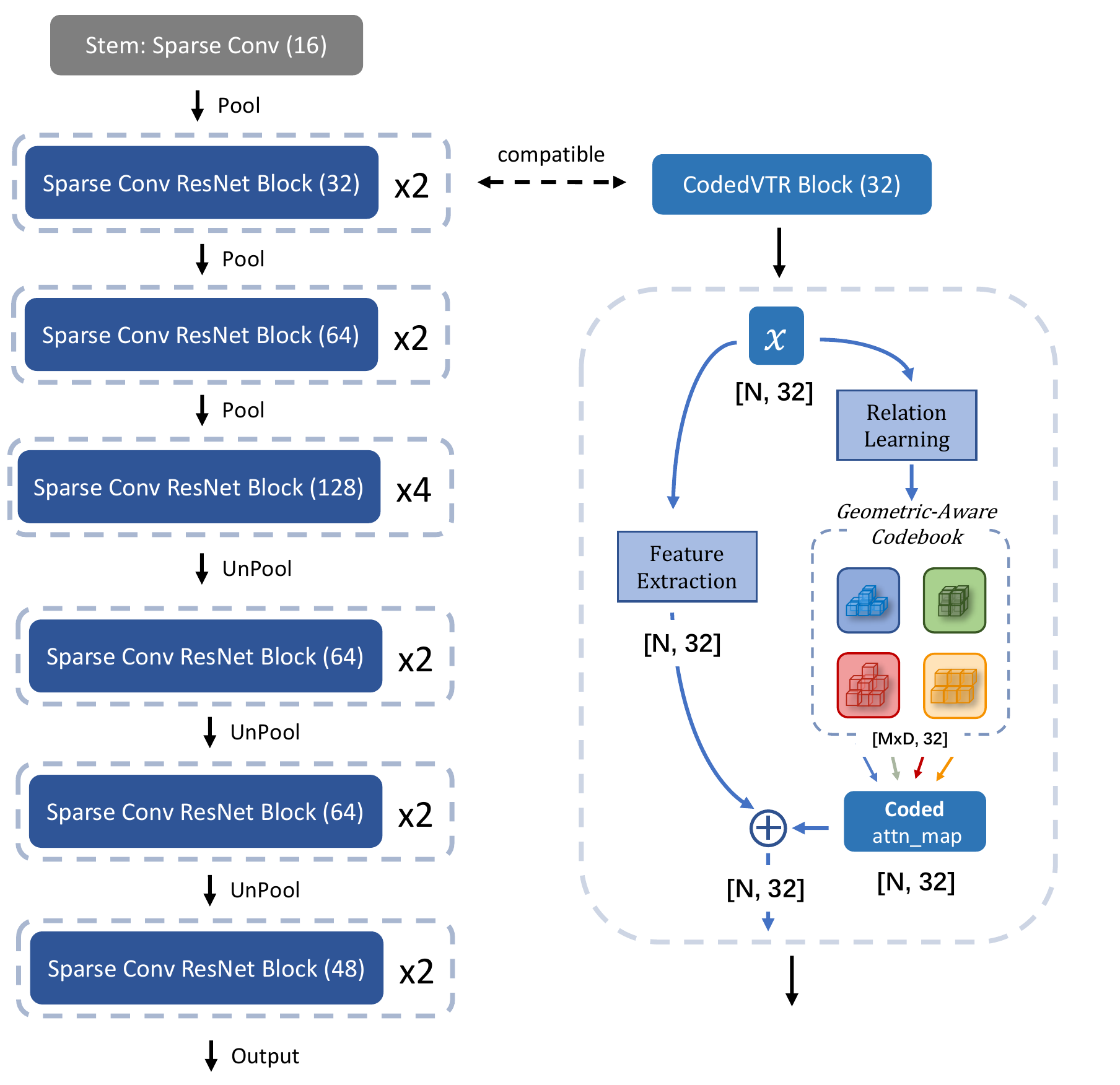}
    \caption{\textbf{The architecture of \method.} The \method shares the backbone architecture like MinkowskiNet~\cite{mink} while replacing the convolution block with \method block.}
    \label{fig:macro-arch}
\end{figure}

\begin{figure}[h]
  \centering
    \includegraphics[width=\linewidth]{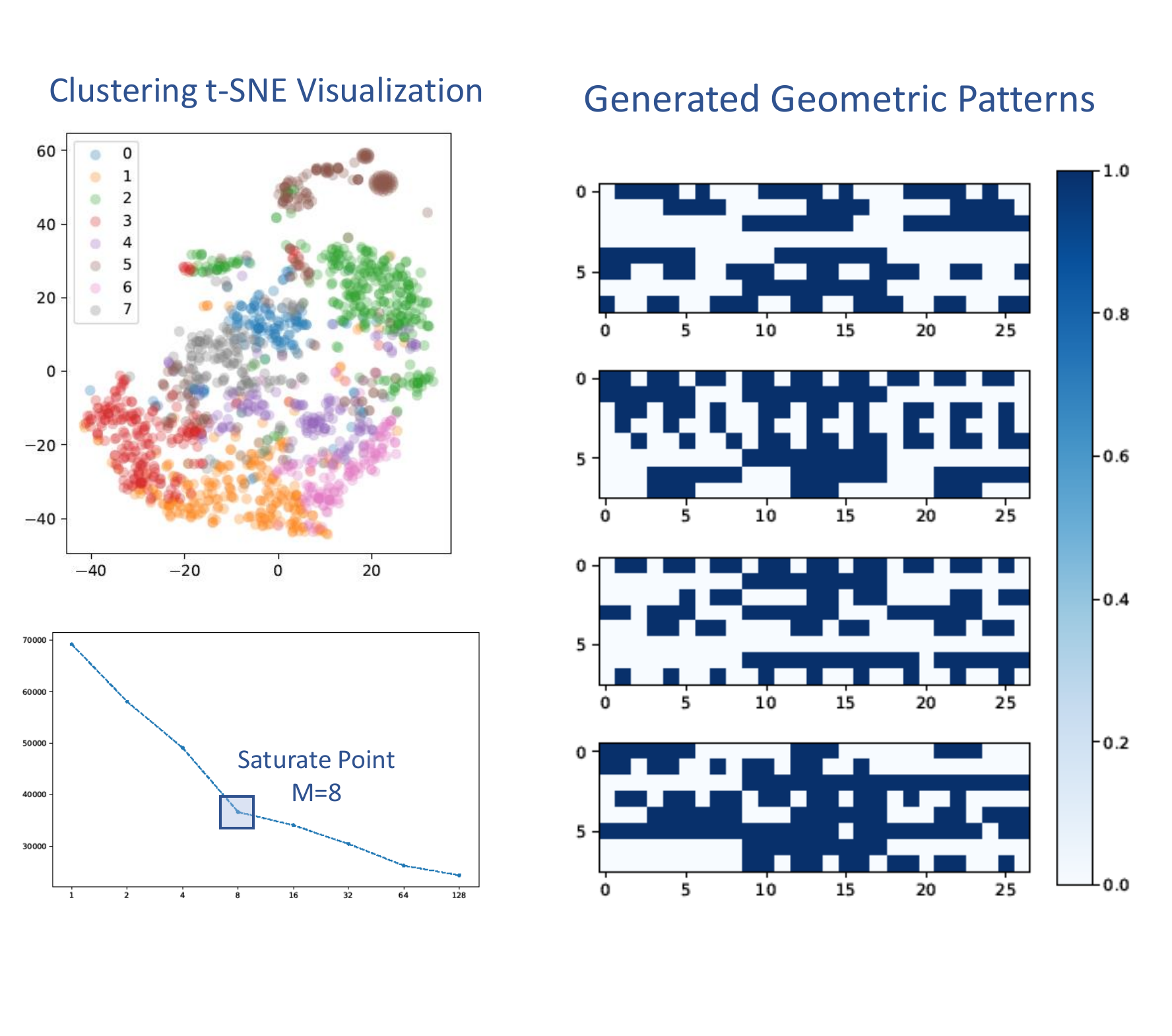}
    \caption{\textbf{The design process of geometric pattern.} The geometric patterns are generated through clustering on the neighbor sparse masks for each dataset, stride and dilation.}
    \label{fig:cluster}
\end{figure}

\section{Design of Geometric Pattern}

In this section, we give a detailed description of the design process of the geometric pattern. We apply clustering on the one-hot neighbor sparse masks for each dataset, stride, and dilation. We randomly sample 10 scenes for each dataset and acquire the neighbor sparse mask in the $3\times3\times3$ region with different strides and dilations. It could be represented with a 27-dimension one-hot sparse mask, where each element represents whether the neighbor location is occupied or not. We apply K-modes clustering~\cite{kmodes} to generate $M$ representative sparse patterns, and the clustering centroids are chosen as the geometric pattern used in geometry-aware attention. We take the dilation=1, stride=2, sparse patterns on semanticKITTI as an example. Fig.~\ref{fig:cluster} visualizes the t-SNE of the clustered sparse patterns. The hyper-parameter $M$ is chosen by investigating the ``saturate point'' of the clustering error, as illustrated in Fig.~\ref{fig:cluster}, we set $M$ as 8.

\begin{figure}[h]
  \centering
    \includegraphics[width=\linewidth]{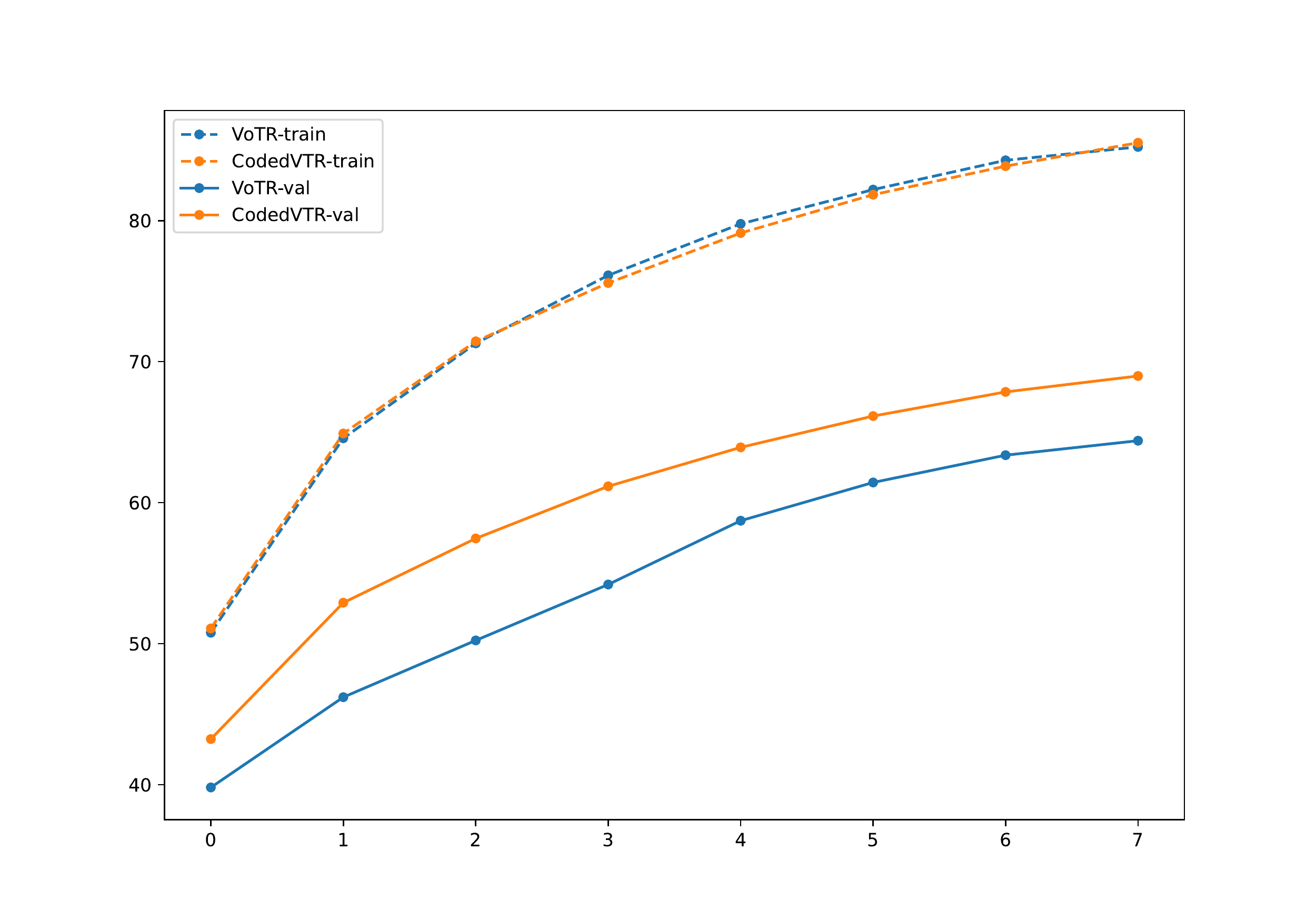}
    \caption{\textbf{Comparison of the generalization for voxel transformer and \method} With similar training accuracy, the \method has superior validation accuracy, proving that it has better generalization ability compared with the original voxel transformer.}
    \label{fig:gene}
\end{figure}

\section{Comparison of Model Generalization}

In this section, we present the analysis of model generalization. We tune the training hyperparameters (weight decay and learning rate) to align the training accuracy for VoTR (voxel transformer) and our \method under similar model capacity. As shown in Fig.~\ref{fig:gene}, the \method has notably higher performance on the validation set, denoting that it has better generalization ability.

\section{Limitations and Future Work}
This section  
discusses possible directions or approaches to improve our proposed \method.
Firstly, when the codebook-based attention has ``hard'' choices, it becomes a ``discretized self-attention'' and  has the potential for better efficiency.
In the future, we might explore techniques to train a discretized version of our \method to achieve better inference-time efficiency. Secondly, using the clustering centroids as the geometric patterns of the codebook elements could be suboptimal. Developing more advanced learning-based methods to acquire them in the training process jointly is an interesting future exploration.

\end{document}